\documentclass[journal]{IEEEtran}

\usepackage{times}
\usepackage{epsfig}
\usepackage{graphicx}
\usepackage{amsmath}
\usepackage{amssymb}
\usepackage{multirow}
\usepackage{url}
\usepackage{lipsum}
\usepackage[dvipsnames]{xcolor}
\usepackage{adjustbox}
\usepackage{algpseudocode}
\algnewcommand\And{\textbf{and} }
\algnewcommand{\Or}{\textbf{or}}
\newcommand{\vars}{\texttt}
\usepackage{subfig}
\DeclareMathSizes{12}{30}{16}{12}
\usepackage{textcomp}
\usepackage{gensymb}
\usepackage[LGRgreek]{mathastext}
\usepackage{mathtools}
\usepackage{upgreek}
\usepackage{environ}
\usepackage{float}
\usepackage[utf8]{inputenc}
\usepackage{todonotes}
\usepackage{xcolor}

\definecolor{codegreen}{rgb}{0,0.6,0}
\definecolor{codeblack}{rgb}{0.0,0.0,0.0}
\definecolor{codepurple}{rgb}{0.58,0,0.82}
\definecolor{backcolour}{rgb}{0.95,0.95,0.92}

\usepackage{listings} [caption={My Caption}]
\lstdefinestyle{mystyle}{
    backgroundcolor=\color{backcolour},   
    commentstyle=\color{codegreen},
    keywordstyle=\color{magenta},
    numberstyle=\tiny\color{codeblack},
    stringstyle=\color{codepurple},
    basicstyle=\ttfamily\footnotesize,
    breakatwhitespace=false,         
    breaklines=true,                 
    captionpos=b,                    
    keepspaces=true,                 
    numbers=left, 
    frame = bottomline,
    numbersep=5pt,                  
    showspaces=false,                
    showstringspaces=false,
    numberblanklines=false,
    showtabs=false,                  
    tabsize=2
}
\usepackage[ruled,vlined]{algorithm2e}

\SetCommentSty{mycommfont}
\SetKwInput{KwInput}{Input}
\SetKwInput{KwOutput}{Output}
\SetKwInput{KwParameters}{Parameters}

\ifCLASSINFOpdf
\else
\fi
\begin{document}
%

\title{On the Generalizability of Motion Models for Road Users in Heterogeneous Shared Traffic Spaces}

%
%
%

\author{
        Fatema T. Johora,~Dongfang Yang,~\IEEEmembership{Student Member,~IEEE,}~Jörg P. Müller,~and~Ümit Özgüner,~\IEEEmembership{Life~Fellow,~IEEE}
    
\thanks{Fatema T. Johora and Jörg P. Müller are from the Department of Informatics, TU Clausthal, Julius-Albert-Str. 4, 38678 Clausthal-Zellerfeld, Germany.}

\thanks{Dongfang Yang, and Ümit Özgüner are from the Department of Electrical and Computer Engineering, The Ohio State University, Columbus, OH 43212 USA.}
\thanks{This work has been submitted to the IEEE for possible publication. Copyright may be transferred without notice, after which this version may no longer be accessible.}
}

%
%

\markboth{Journal of \LaTeX\ Class Files,~Vol.~xx, No.~x, Month~Year}%
{Shell \MakeLowercase{\textit{et al.}}: Bare Demo of IEEEtran.cls for IEEE Journals}
%



\maketitle

\begin{abstract}
Modeling mixed-traffic motion and interactions is crucial to assess safety, efficiency, and feasibility of future urban areas. The lack of traffic regulations, diverse transport modes, and the dynamic nature of mixed-traffic zones like shared spaces make realistic modeling of such environments challenging. This paper focuses on the generalizability of the motion model, i.e., its ability to generate realistic behavior in different environmental settings, an aspect which is lacking in existing works. Specifically, our first contribution is a novel and systematic process of formulating general motion models and application of this process is to extend our Game-Theoretic Social Force Model (GSFM) towards a general model for generating a large variety of motion behaviors of pedestrians and cars from different shared spaces. Our second contribution is to consider different motion patterns of pedestrians by calibrating motion-related features of individual pedestrian and clustering them into groups. We analyze two clustering approaches. The calibration and evaluation of our model are performed on three different shared space data sets. The results indicate that our model can realistically simulate a wide range of motion behaviors and interaction scenarios, and that adding different motion patterns of pedestrians into our model improves its performance.

\end{abstract}


%
\IEEEpeerreviewmaketitle

\section{Introduction}
\label{sec:intro}
\setlength{\parindent}{2ex}
\IEEEPARstart{S}{hared} space design principles~\cite{clarke2006shared} have been drawing significant attention in recent years, as an alternative to traditional regulated traffic designs. In shared spaces, heterogeneous road users such as pedestrians, cars, bicycles share the same space. Unlike traditional traffic environments, in shared spaces, there are no or very few road signs, signals, and markings; this causes frequent direct interactions among road users to coordinate their trajectories. 

There is an ongoing debate on the safeness of shared spaces; while some studies state that the lack of explicit traffic regulations makes road users more safety-conscious and may lead to fewer road accidents \cite{hamilton2008shared, clarke2006shared, kaparias2012analysing},  others (\cite{clayden2006improving,jenks1983residential}) argue the lack of acceptance and understanding of the concept can compromise safety in shared spaces. 
Notwithstanding this debate, traditional road designs have been replaced by shared spaces in a growing number of urban areas; some examples are the Laweiplein intersection in Drachten, Skvallertorget in Norrk{\"o}ping, and Kensington High Street in London \cite{hamilton2008shared}. 

Yet, the lack of explicit rules makes it essential to investigate the safety issues in shared spaces. Modeling and simulation shared spaces by analyzing and reproducing the motion behaviors of road users including their interactions is crucial to assess and optimize such spaces during the planning phase. Realistic simulation models can also form a safe basis for autonomous cars to learn how to interact with other road users.

Interpreting and modeling mixed-traffic interactions pose challenging problems; an interaction can be a simple reaction or a result of complex human decision-making processes, i.e., modifying speed or direction by predicting other road users' behavior, or communicating with them \cite{rasouli2019autonomous}. Moreover, how one interacts with others is dependent on many factors like their transport mode, current situation, road structures and conditions, social norms (culture), and many individual factors (e.g. age, gender, or time pressure \cite{kaparias2012analysing}). 

To the best of our knowledge, so far, there are not many works on modeling and simulation of shared spaces. We observe mostly two different state-of-the-art approaches: (1) physics-based models, mainly the social force model (SFM) of pedestrian dynamics \cite{helbing1995social} including numerous extensions adding, e.g., new forces, decision-theoretic concepts, or rule-based constraints, to describe different types of actors such as cars \cite{schonauer2017microscopic,anvari2015modelling} or bicycles \cite{rinke2017multi}; and (2) cellular Automata (CA) models \cite{lan2005inhomogeneous,zhang2007modeling,bandini2017collision}, which are mainly used for modeling mixed-traffic flows in settings with explicit traffic regulations, unlike most shared spaces.
 
Although these approaches perform well for single bilateral conflicts (i.e., for any point in time, a road user can only handle a single explicit conflict with one other user), they fail in representing multiple conflicts among heterogeneous road users and groups, which are very common in shared spaces. Hence, in our previous works, we integrated SFM with a game-theoretic model to address both bilateral and multiple conflicts among pedestrians and cars \cite{johora2018modeling,ahmed2019investigating}. In this paper, we describe \textit{conflict} as ``an observable situation in which two or more road users approach each other in time and space to such an extent that there is a risk of collision if their movements remain unchanged'' as specified in \cite{gettman2003surrogate}; here, we use the terms \textit{conflict} and \textit{interaction} interchangeably.

In the literature, motion models do not adequately consider the differences in road users' behaviors induced by differing environmental settings. These models are usually calibrated and validated using scenarios from a single shared space environment. In \cite{johora2020zone}, we took a first step to address this gap by proposing the concept of zone-specific motion behaviors for pedestrians and cars, considering road and intersection zones. In \cite{johoraTL2020}, we evaluated the transferability\footnote{We use the terms \textit{transferability} and \textit{generalizability} interchangeably} of our existing model by modeling scenarios that differ from the one used in \cite{johora2020zone} in terms of traffic conditions, spatial layout and social norms. Subsequent results show that our model can suitably replicate the motion of pedestrians and cars from the new scenarios.

In this paper, we delve further into this direction by proposing a conceptually systematic and simple process of modeling general motion models and output a moderate version of a general motion model for pedestrians and cars, by following our proposed modeling process. A general model should be able to reproduce a large variety of motion behaviors of heterogeneous road users ranging from simple free-flow motions to resulted-motions from complex interactions and transferable to new environments with minimal time and effort. The differences between the current work and our previous work (\cite{johoraTL2020}) in terms of model transferability are:
(1) In this paper, we build a general model to capture motion behaviors from three data sets with incremental integration of new motion behaviors, and a well-defined and largely automated calibration process to adapt model parameters to the target environment. Whereas in \cite{johoraTL2020}, as we did not have any specific process to generate a general motion model, to adapt to the new environment, we had to analyze, consider and explicitly change our model parameters and methods based on the social norms of that new environment, which resulted in different versions of our model, i.e. each version for each different environment. (2) In the current work, the transferability of our model is evaluated using the DUT and HBS data sets as in \cite{johoraTL2020} and also by a new data set (CITR) that contains unique conflict scenarios than the other two data sets (see Section \ref{se:dataset_and_interaction}).

We further introduce heterogeneity in pedestrian motion by recognizing different motion patterns, by calibrating individual motion characteristics (e.g., sensitivity when interacting with others) and clustering them into different groups \footnote{In this paper, the keyword \textbf{group} is used to represent a set of pedestrians with a similar motion pattern, not the social group, e.g., family members.} (see Section \ref{se:calibration}). The contributions of this paper are:
\begin{itemize}
\item We propose a systematic process to formulate a general motion model.
\item We propose a motion model for pedestrians and cars, which can simulate a large variety of conflict scenarios among road users and evaluate the generalizability of our model by using three different shared space data sets. The results of our evaluation process indicate that our model achieves satisfactory performance for each data set.
\item We present a methodology to recognize and model different motion patterns of pedestrians from real-world data sets. To do so, we investigate several approaches to cluster pedestrians with similar motion patterns into groups. Our evaluation results show that the heterogeneity in pedestrians motion improves the model performance.
\end{itemize} 
Following a review of previous research in Section
\ref{sec:related_work}, we propose the formulation of a general model for movement modeling of heterogeneous road users in Section \ref{se:modeling_process}. We illustrate the examined data sets and the architecture of our Game-Theoretic Social Force Model (GSFM) in Section~\ref{se:dataset_and_interaction} and Section~\ref{se:model}, respectively.
Section \ref{se:calibration} explains the calibration methodology and recognition of different walking styles of pedestrians. In Section \ref{se:evaluation}, we describe how we evaluate model performance and discuss the results. We conclude by outlining future research venues. 

\section{Related Works}
\label{sec:related_work}
\setlength{\parindent}{2ex}
Existing mixed-traffic motion models are mostly built based on rule-based models (e.g. Cellular Automata (CA) \cite{zhang2007modeling}), or physics-based models, most preeminently the Social Force Model (SFM) \cite{helbing1995social}.

CA models describe road users motion behavior by a set of state transforming rules in a discrete environment. They have been used to model motion behaviors of a set of homogeneous road users, e.g., pedestrians \cite{burstedde2001simulation,bandini2017approach}, cars \cite{nagel1992cellular,chai2015fuzzy} and there are also few works describing mixed-traffic motion,  e.g.,~\cite{zhang2007modeling} who study interactions among pedestrians and cars at crosswalks, \cite{lan2005inhomogeneous} who model car-following and lane-changing actions of cars and motorcycles, or \cite{chen2018evaluating} who study bicycle-to-vehicle interactions and its impact on traffic delay. 

In the classical SFM, introduced in \cite{helbing1995social}, the movement of a pedestrian is represented by differential equations comprising a set of simple attractive, and repulsive forces from other pedestrians and static obstacles that he/she experiences at a specific place and time. Even though SFM was initially modeled for pedestrian dynamics \cite{chen2018social,asano2010microscopic,johora2017dynamic}, many studies extended it for modeling other types of road users. For example, \cite{yang2018social,zeng2014modified} who include vehicles, considering their impact on pedestrians as separate forces; in \cite{anvari2015modelling}, Anvari et al.~add new forces and rule-based constraints to handle short-range and long-range conflicts among pedestrians and cars. In \cite{rinke2017multi}, SFM is combined with long-range collision avoidance mechanisms to model  motion behaviors of pedestrians, vehicles and bicycles.

Both CA-based and SFM-based models can represent simple situations well. However,  game-theoretic or probabilistic models are more suitable for complex scenarios where road users must choose an action among many alternatives to handle a given situation \cite{helbing1995social}. In \cite{pascucci2017discrete}, in case of complex interactions, road users' choice of action is modeled by a logit model, based on available data but without considering what other users might do. In \cite{fujii2017agent}, Fujii et al.~used a discrete choice model to illustrate decision making while in pedestrian interactions.
Game-theoretic models have often been applied to interpret human decision-making processes, also in traffic situations. Some examples are the application of non-cooperative games to illustrate merging-give way interaction among vehicles (\cite{kita1999merging}), pedestrian-to-car interaction in shared spaces (\cite{schonauer2017microscopic}), bicyclist-to-car interaction at zebra crossings (\cite{bjornskau2017zebra}), or analyze the difference of cyclist/pedestrian interaction with human-driven or autonomous vehicles in \cite{michieli2018game}. In \cite{lutteken2016using}, lane-changing behaviors of cars are modeled using a cooperative game where cars cooperate with each other for collective reward. Whereas, in a non-cooperative game, each player makes decisions by predicting others' decisions, which is very similar to what real-world road users often do \cite{bjornskau2017zebra}. 

Although there are several works on modeling motion behavior of road users, only a very few studies consider different motion patterns for individual road user types \cite{kabtoul2020towards, alahi2017learning, yu2014multiagent}. Kabtoul et al. \cite{kabtoul2020towards} manually annotates several predefined pedestrian types based on willingness to give way to a car. 
Alahi et al.~\cite{alahi2017learning} obtain different movement styles for pedestrians by learning collision avoidance parameters of individual pedestrians and clustering them into groups using the k-means clustering. Their model is restricted to pedestrian-only scenarios. In \cite{yu2014multiagent}, the authors classified pedestrians into groups based on their age range and gender and assigned individual speed profiles to each group. These speed profiles are collected from the literature instead of real-world data sets. 

Existing closed-source commercial (e.g., AIMSUN \cite{casas2010traffic} or VISSIM \cite{fellendorf2010microscopic}) and open-source (SUMO \cite{behrisch2011sumo}) simulators are somewhat capable of modeling and simulating mixed-traffic at a microscopic level. However, open-source simulators like SUMO have limited means for modeling interaction between heterogeneous road users. To address this issue, some studies combined SUMO with agent frameworks such as JADE (\cite{soares2013agent}) or JASON (\cite{Fiosins+2016arts}); however, adding new environmental features or define new modalities in such models is difficult. Also, SUMO lacks flexibility regarding lane and vehicle geometries, which is restrictive for shared spaces.


\section{Modeling Process}
\label{se:modeling_process}
A general motion model should be able to reproduce realistic motion behaviors of road users in different environmental settings in terms of road structures, culture or norm, types of road users, and types of interactions and to adapt to new environments with less time and effort, which make generating such models very challenging. 

\begin{figure}[htbp]
	\centering
	\vspace{-2mm}
	\includegraphics{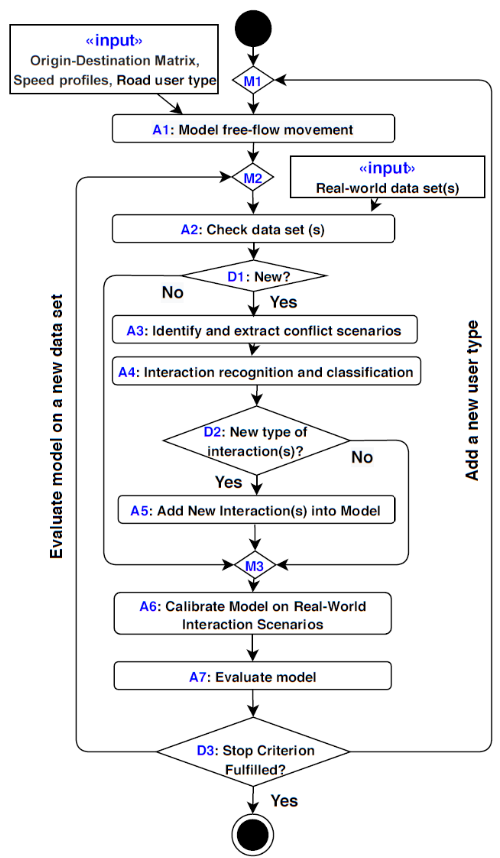}
	\caption[Formulation of a general motion model for mixed-traffic environments]{Formulation of a general motion model for mixed-traffic environments.}
	\label{fig:generic_model}
	\vspace{-3.5mm}
\end{figure}

We propose a systematic process to construct a general motion model in Figure \ref{fig:generic_model}. Here, D, A, and M represents the decision, action and merge nodes respectively. The process starts with modeling the free-flow movements of road users (\textbf{A1}) with their type and origin, destination, and speed profiles as \textit{input}. The next step is to analyze and model interactions among road users. To do so, one can collect and explore a real-world traffic data set (\textbf{A2}) to identify and extract conflict scenarios between two or more road users (\textbf{A3}) to recognize and classify the interactions among the road users (\textbf{A4}) and then model these interactions (\textbf{A5}). Finally, the model needs to be calibrated (\textbf{A6}) and evaluated (\textbf{A7}) both quantitatively (minimize the difference between real and generated trajectories) and qualitatively (reproduce realistic behaviors) by using these extracted conflict scenarios. However, generating a general motion model is a continuous process which requires testing the model with new data sets, i.e., new environments and also adding new modalities.
As shown in Figure \ref{fig:generic_model}, to evaluate the model performance on a new (\textbf{D1}) data set, it is necessary to check (\textbf{D2}) if there are any new kind of interaction(s), if yes, then this interaction(s) needs to be integrated (\textbf{A5}) into the model. Next, the calibration of all parameters (including the new ones) and the model evaluation on each data set is required. To add a new user type (\textbf{M1}) e.g., integrating vehicle in the pedestrian-only motion model, one needs to go through all the steps in Figure \ref{fig:generic_model}. This iterative process of modeling continues until a stopping criterion, such as a certain level of accuracy in realistic trajectory modeling, has been reached. The stopping criterion is application dependent. 

In this paper, we use this process to \textit{output} a moderate version of a general model for generating realistic trajectories of pedestrians and cars in different shared spaces, using the HBS, DUT and CITR data sets. Our way of recognizing and classification of interactions (\textbf{A4}), modeling these interactions (\textbf{A5}), the calibration (\textbf{A6}) and evaluation (\textbf{A7}) of the model are discussed in Section \ref{sec:interactionClassification}, \ref{se:model}, \ref{se:calibration} and \ref{se:evaluation}, respectively.

\section{Data Sets and Interaction Classifications}
\label{se:dataset_and_interaction}
\setlength{\parindent}{2ex}
\subsection{Data Sets}
\label{sec:datasets}

\begin{figure}[htbp]
   \centering
   \includegraphics[width=3.45in,height=2.15in]{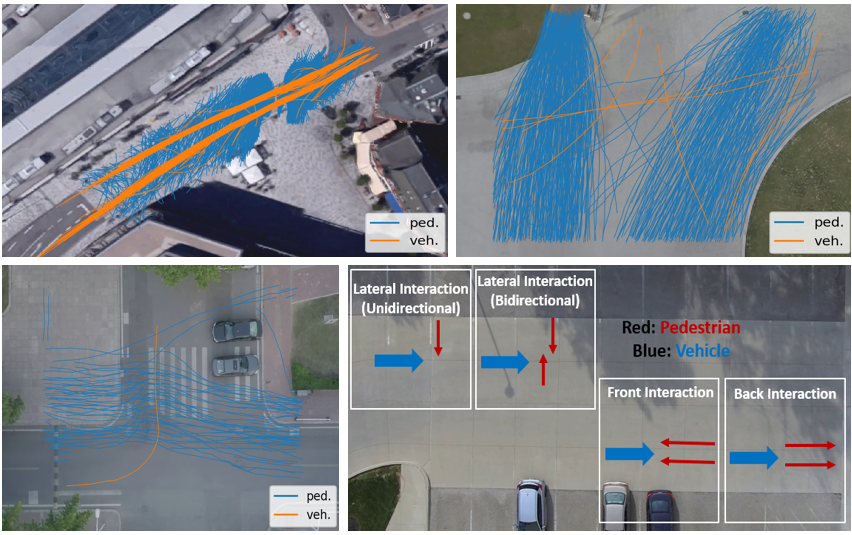}
   \caption[Mixed-traffic data sets]{The spatial layout of three shared space environments; the top-left sub-figure visualizes the shared street from HBS, the top-right sub-plot shows the roundabout from DUT, the bottom-left sub-plot depicts the intersection from DUT and the bottom-right sub-figure shows interactions from CITR.}
  \label{fig:dataset}
  \vspace{-3.5mm}
\end{figure}

We have been developing a motion model of pedestrians and cars, named Game-Theoretic Social Force Model (GSFM) \cite{johora2018modeling, johora2020zone, Johora2020agent}, mainly based on the scenarios manually extracted from a street-like shared space environment in Hamburg, Germany (HBS). In this paper, to move towards a general model, we evaluate our model on two other data sets which are different from the HBS data set in terms of spatial structures, types of interactions, and the number of road users. These data sets are the DUT data set from Dalian University of Technology campus in China and the CITR data set from the Ohio State University campus in the USA. All three data sets are visualized in Figure \ref{fig:dataset}, and their details are given below:
\begin{itemize}
    \item \textbf{HBS} \cite{rinke2017multi}: The HBS data set collected from a street with pedestrian crossing from both sides. It contains both bilateral and multilateral interactions among pedestrians and cars, with or without car following interactions. We extracted 103 such scenarios from HBS.
    \item \textbf{DUT} \cite{yang2019top}: The DUT data set contains trajectories of pedestrians and cars from a roundabout and an intersection. It comprises of car-to-crowd lateral interactions; most scenarios extracted from DUT have a large number of pedestrians compared to the HBS and CITR scenarios.
    \item \textbf{CITR} \cite{yang2019top}: CITR is an experimentally designed data set, collected from a university parking lot. It contains several lateral, front, and back interactions among pedestrians and cars.
\end{itemize}
Here as shown in the bottom-right sub-figure of Figure \ref{fig:dataset}, lateral interaction indicates the situation where pedestrian(s) cross from in front or behind the car. Front interaction is the face-to-face interaction, and in back interaction scenario, car drives behind the pedestrian(s). There are also observable differences in these data sets which can be interpreted as cultural differences. For example, in the DUT data set, road users maintain less inter-distance (i.e., safety distance) compared to the HBS and CITR data sets (see Section \ref{se:calibration}).
In all three data sets, an agent's position at each time step (i.e., 0.5 s) is given as a 2D vector in the pixel coordinate system, and they also contain the pixel-to-meter conversion scales. Table \ref{tab:data_statistics} summarizes the number of scenarios and individuals involved.
\begin{table}[H]
    \centering
    \caption{Statistics of Datasets}
    \begin{tabular}{|c|c|c|c|c|}
        \hline
        Data set & \# of Scenarios & \# of Pedestrians & \# of Cars & Time step   \\ \hline 
        HBS     & 103 & 206 & 126 & 0.5s\\ \hline
        CITR    & 26  & 208 & 26 & 0.5s\\ \hline 
        DUT     & 30  & 607 & 39 & 0.5s\\ \hline 
    \end{tabular}
    \label{tab:data_statistics}
    	\vspace{-2.5mm}
\end{table}

\subsection{Interaction Classification}
\label{sec:interactionClassification}
In our previous works \cite{johora2018modeling, Johora2020agent}, we classified road users interactions broadly into two categories based on Helbing's classification of road agents' behavior \cite{helbing1995social} and the observation of the shared space video data (mostly HBS): \textit{simple interaction} (percept $\rightarrow$ act) and \textit{complex interaction} (percept $\rightarrow$ choose an action from different alternatives $\rightarrow$ act). These interactions can also be sub-categorized based on the number and types of road users involved: \textit{simple interaction} contains car-following, pedestrian-to-pedestrian, and pedestrian(s)-to-car reactive interactions and \textit{complex interaction} includes pedestrian(s)-to-cars, pedestrians-to-car and car-to-car interactions. We note that complex car-to-car interaction is not included in this paper.

As mentioned earlier, in this paper, we are still focusing on pedestrians and cars, but we aim to evaluate the performance of our model on the DUT and CITR data sets. According to the process proposed in Figure \ref{fig:generic_model}, we analyze these two data sets and detect the following new types of interactions:
\begin{itemize}
    \item Unlike HBS, in the DUT data set, sometimes, cars somewhat deviate from their trajectory as a result of reactive interaction with pedestrians. Mostly because of the environment structure in DUT, i.e., more free space for motion of cars. 
    \item As already discussed in Section \ref{sec:datasets}, the CITR data set \cite{yang2019top} contains front and back interactions among pedestrians and cars, which are not observed in the HBS or DUT data sets \cite{rinke2017multi}.
\end{itemize}  
How we model these interactions, including  integration of new interaction types, is described in Section \ref{se:model}.

\section{Agent-Based Simulation Model}
\label{se:model}
\setlength{\parindent}{2ex}
\begin{figure}[htbp]
	\vspace{-2mm}
	\centering
	\includegraphics[width=3in,height=2.8in]{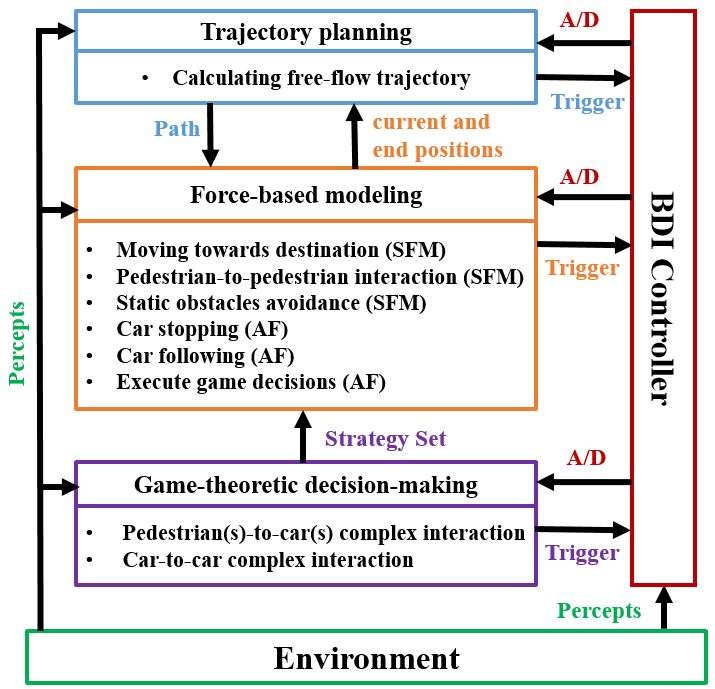}
	\caption[Conceptual model of pedestrians and cars motion behaviors]{Conceptual model of pedestrians and cars motion behaviors. Here, \textbf{AF} denotes the added force to classical SFM and \textbf{A/D} signifies activation/deactivation of a module.}
	\label{fig:architecture}
	\vspace{-3.5mm}
\end{figure}

We pursue an agent-based model, GSFM, to represent the motion behaviors of pedestrians and cars, initially described in \cite{johora2018modeling}. Here, we give an overview of the architecture of GSFM, visualized in Figure~\ref{fig:architecture}. In GSFM, each road users is modeled as an individual \textit{agent} and their movements are conducted in three interacting modules, namely, \textit{trajectory planning}, \textit{force-based modeling}, and \textit{game-theoretic decision-making}. Each of this module has individual roles. GSFM is implemented on a BDI (\textbf{B}elief, \textbf{D}esire, \textbf{I}ntention) platform, LightJason \cite{aschermann2016lightjason}, which permits flexible design and explanation of the control flow of GSFM through its three modules. Based on current situation, the \textit{BDI controller} activates the relevant module, which then informs the controller on completion of its task.

\textbf{The trajectory planning module} computes the free-flow trajectories for all agents by only considering static obstacles (e.g. boundaries, or trees) in the environment. For individuals trajectory planning, we transform the simulation environment into a visibility graph \cite{koefoed2012representations}, add their origin and destination positions into the graph and perform the A* algorithm \cite{millington2009artificial}.

\textbf{The force-based module} governs the actual execution of an agent's physical movement and also captures the simple interactions between agents by using and extending the SFM. To model the driving force of agents towards their destination ({$\vec{D}_{i}^o$}), the repulsive force from the static obstacles ($\vec{I}_{i W}$) and other agents ($\vec{I}_{i j}$), we use the classical SFM. Here, $\vec{D}_{i}^o$ = $\frac{\vec{v^*}_i(t) - \vec{v}_i(t)}{\tau}$ for a relaxation time $\tau$ and $\vec{v^*}_i(t)$ and $\vec{v}_i(t)$ denote the desired and current velocities of $i$, $\vec{I}_{i j}$ = $V_{i j}^o\exp\bigg[\frac{ r_i + r_j - \vec{d}_{i j}(t)}\sigma\bigg] \hat{n}_{i j} F_{i j}$ 
and 
$\vec{I}_{i W}$ =  $U_{i W}^o\exp\bigg[\frac{ r_i - \vec{d}_{i W}(t)}{\gamma}\bigg] \hat{n}_{i W}$, where $V_{i j}^o$ and $U_{i W}^o$ symbolize the interaction strengths, and $\sigma$ and $\gamma$ are the range of these repulsive interactions, $\vec{d}_{i j}(t)$ and $\vec{d}_{i W}(t)$ are the distances from $i$ to $j$, or $i$ to $W$ at a specific time, $\hat{n}_{i j}$ and $\hat{n}_{i W}$ indicate the normalized vectors. $F_{i j}$ = $\lambda + (1 - \lambda)\frac{1+ \cos{\upvarphi_{i j}}}{2}$ describes the fact that human are mostly affected by the objects which are within their field of view \cite{johansson2008specification}. Here, $\lambda$ stands for the strength of interactions from behind and $\upvarphi_{i j}$ symbolizes the angle between $i$ and $j$.
Additionally, we extend SFM to represent car following interaction ($\vec{I}_\text{follow}$) and pedestrian-to-car reactive interaction ($\vec{I}_\text{stop}$). 
If $\vec{d}_{i j}(t)$ $\geq$ $D_{min}$, $\vec{I}_\text{follow}$ = $\hat{n}_{p_i x_i (t)}$, i.e., $i$ continues moving towards $p_i = x_i (t) + \hat{v}_j(t) * D_{min}$, otherwise, $i$ decelerates. Here, $D_{min}$ is the minimum vehicle distance, $\hat{v}_j(t)$ is the normalized velocity of j, and $\vec{d}_{i j}(t)$ denotes the distance between $i$ and $j$ (the leader car).
$\vec{I}_\text{stop}$ emerges only if pedestrian(s) have already begun walking in-front of the car. Then the car decelerates to let the pedestrian(s) proceed. This module also executes the decisions computed in the game module $\vec{I}_{\text{game}}$.

As discussed in Section \ref{sec:interactionClassification}, the CITR data set contains two new types of interaction, namely, the front and back interaction between pedestrian ($i$) and vehicle ($j$). We incorporate these two interactions to our model as a single type, i.e., longitudinal interaction, $\vec{I}_{long}$ and  following:

If $\vec{d}_{i j}(t) < D_{min}^{long}$
and ($C_1$ or ($C_2$ and $C_3$) ), we add a temporary goal $p_{i} = \vec{x}_i (t) + R_f$ for the respective pedestrian, where $C_1$, $C_2$, and $C_3$ are symbolized in Eq. \eqref{eq:conditions} with $g = \vec{e}_i\cdot\vec{e}_j$, i.e., the dot product of the direction vectors of $i$ and $j$, and $R_f$ is the rotation of $f = \vec{e}_j * c$ using rotation theory in Eq. \eqref{eq:rotation} \cite{weisstein2003rotation} and the calculation of $c$ and $\theta$ are given in Eq. \eqref{eq:cd} and Eq. \eqref{eq:theta} respectively. Thus, $\vec{I}_{long}$ = $\hat{n}_{p_i} x_i(t)$, i.e., $i$ continues moving towards $p_i$ to avoid conflict.
 \vspace{-1mm}
\begin{equation}
\begin{aligned}
 C_1 = \theta_{\vec{e}_j\hat{n}_{ji}} < 2\degree \hspace{0.1cm} or \hspace{0.1cm} \theta_{\vec{e}_j\hat{n}_{ji}} > 358\degree
 \\
 C_2 = g \geq 0.99 \hspace{0.1cm} or \hspace{0.1cm} g \leq -0.99
 \\
 C_3 = \theta_{\vec{e}_j\hat{n}_{ji}} \geq 348\degree \hspace{0.1cm} or \hspace{0.1cm} \theta_{\vec{e}_j\hat{n}_{ji}} \leq 12\degree
 \end{aligned}
\label{eq:conditions}
\end{equation}

\begin{equation}
  \begin{aligned}
  f_{x_2} = \cos\theta f_x - \sin \theta f_y  \\
  f_{y_2} = \sin\theta f_x + \cos\theta f_y
  \end{aligned}
 \label{eq:rotation}
\end{equation}
\begin{equation}
\theta =
    \begin{dcases}
     90\degree, & if \hspace{0.05cm} \theta_{\vec{e}_j\hat{n}_{ji}} \geq 348\degree \\
     180\degree, & \text{otherwise}
    \end{dcases}
    \label{eq:theta}
\end{equation}

\begin{equation}
  b =
  \begin{dcases}
     1, & if \hspace{0.05cm} g \leq -0.99 \\
     1.5, & \text{otherwise}
   \end{dcases}
   \\
   c =
  \begin{dcases}
     3*b, & if \hspace{0.05cm} \theta_{\vec{e}_j\hat{n}_{ji}} \geq 348\degree \\
     2.2*b, & \text{otherwise}
  \end{dcases}
  \label{eq:cd}
\end{equation}
In this paper, $D_{min}^{long}$ is set to 10 m. Deviation of cars due to reactive interaction with pedestrian in the DUT scenarios is addressed by $\vec{I}_{i j}$, i.e., the SFM repulsive force. 

\begin{figure*}[!htbp]
	\centering
	\subfloat[Pedestrian-to-Car Interaction]{\includegraphics[width=1.7in, height = 1.15in]{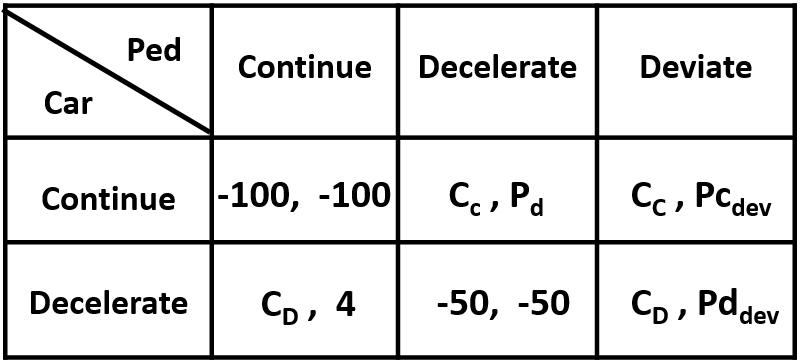}}
	\hfil
	\subfloat[Impacts of Situation Dynamics]{\includegraphics[width=5.3in, height = 1.15in]{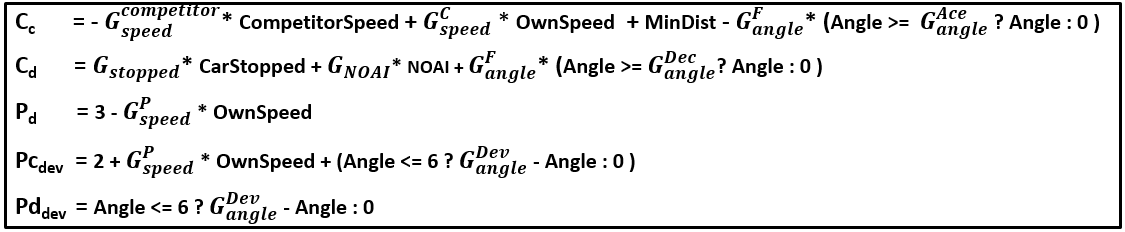}}
	\caption{The complete payoff matrices for pedestrian-to-car interactions.}
	\label{gamematrix}
    \vspace{-1.5em}
\end{figure*}

\textbf{The game-theoretic module} controls the complex interactions among agents, e.g. pedestrians-to-cars interaction, using Stackelberg game, i.e., a sequential leader-follower game. In a Stackelberg game, first, the leader decides on a strategy that maximizes its utility by predicting all possible reactions of followers and then, the follower reacts by choosing its best response \cite{schonauer2017microscopic}. The game is solved by finding the sub-game perfect Nash equilibrium (SPNE) i.e., the optimal strategy pair. The Eq.~\eqref{eq:SPNE} and Eq.~\eqref{eq:Bs_f} depict the SPNE and the best response of the follower, respectively. Here, $s_l$, $s_f$, $u_l$, $u_f$ and $S_l$, $S_f$ are the leader's and follower's strategies, utilities of the corresponding strategies and their strategy sets, respectively.
\begin{equation}
\text{SPNE} = \{s_l\in S_l | max(u_l(s_l, Bs_f(s_l)))\}, \, \forall s_l\in S_l.
\label{eq:SPNE}
\end{equation}
\begin{equation}
Bs_f(s_l) = \{s_f\in S_f|max(u_f(s_f|s_l))\}.
\label{eq:Bs_f}
\end{equation}
An individual game manages each complex interaction, and the games are independent on each other. In each game, the number of leaders is fixed to one but the followers can be more. We perform separate experiments with randomly chosen leader, the faster agent as leader (i.e., the car), and pedestrian as a leader. The result suggests that and the faster agent as leader is the best choice. However, if the scenario includes more than one car (e.g., pedestrian-to-cars interaction), then the one who recognizes the conflict first is considered as the leader. To calculate the payoff matrix of the game, as shown in Figure \ref{gamematrix}, first, all actions of the players are ordinally valued, assuming that they prefer to reach their destination safely and promptly. Then, to express situation dynamics, we select several features by analyzing real-world situations and perform a backward elimination process on the selected features to get the most relevant ones. Let, $i$ be an agent which interacts with another agent $j$; then the relevant features are the following: 

\begin{itemize}
	\item \textbf{NOAI}: the number of active interactions of \textit{i} as a car.
	\item \textbf{CarStopped}: has value $1$ if \textit{i} (as a car) already stopping to give way to another agent \textit{j'}, otherwise 0. 
	\item \textbf{MinDist}: has value $G^{min}_{dis}$- distance(\textit{i}, \textit{j}), if distance(\textit{i}, \textit{j}) $<$ $G^{min}_{dis}$; its difficult to stop for car \textit{i}, otherwise 0.
	\item\textbf{CompetitorSpeed}: has value $1$, if current speed of \textit{j}, $S_{current}$ $<$ $S_{normal}$, otherwise 0. 
	\item\textbf{OwnSpeed}: 
	\[ \begin{dcases}
S_{current}, & if \hspace{0.12cm} \textit{i} \hspace{0.12cm}is \hspace{0.12cm}a \hspace{0.12cm}car \\
	1, & if \hspace{0.12cm} \textit{i} \hspace{0.12cm}is \hspace{0.12cm}a \hspace{0.12cm}pedestrian \hspace{0.12cm}and\hspace{0.12cm} S_{current} > S_{high}  \\
	0, & \text{otherwise}
	\end{dcases}\]
	\item \textbf{Angle}: 
\[
\begin{dcases}
    8,& if (\theta_{\vec{e}_j\hat{n}_{ij}}< 16\degree \hspace{0.12cm}and\hspace{0.12cm} \geq 0\degree) \hspace{0.03cm} \Or \hspace{0.03cm} \theta_{\vec{e}_j\hat{n}_{ij}}>344\\
    7,&
    \parbox{5.75cm}{if$ \hspace{0.03cm} (\theta_{\vec{e}_j\hat{n}_{ij}}\leq 42\degree \hspace{0.1cm}and\hspace{0.1cm}\geq 16\degree) \hspace{0.1cm} \Or \hspace{0.1cm}  (\theta_{\vec{e}_j\hat{n}_{ij}}\leq 344\degree \hspace{0.25cm}and\hspace{0.12cm} \geq 318\degree)$}
    \\
    6,& \parbox{5.75cm}{if$ \hspace{0.03cm} (\theta_{\vec{e}_j\hat{n}_{ij}}\leq 65\degree \hspace{0.1cm}and\hspace{0.1cm} > 42\degree) \hspace{0.1cm} \Or \hspace{0.1cm} \notag (\theta_{\vec{e}_j\hat{n}_{ij}}< 318\degree \hspace{0.1cm}and\hspace{0.1cm} \geq 295\degree)$}\\
    5,& \parbox{5.75cm}{if$ \hspace{0.03cm} \hspace{0.02cm} (\theta_{\vec{e}_j\hat{n}_{ij}}\leq 90\degree \hspace{0.1cm}and\hspace{0.1cm} > 65\degree) \hspace{0.02cm} \Or \hspace{0.02cm} \notag (\theta_{\vec{e}_j\hat{n}_{ij}}< 295\degree \hspace{0.1cm}and\hspace{0.1cm} \geq 270\degree)$}\\
    1,              & \text{otherwise}
\end{dcases}
\]
\end{itemize}
During game playing, \textit{Continue}, \textit{Decelerate} and \textit{Deviate} (only for pedestrian) are the viable actions for road users. Execution of these actions are performed in the force-based module.
\begin{itemize}
	\item Continue: Any pedestrian \textit{i} crosses a car \textit{j} from the point $p_i = {x}_j(t) + S_{A} * \overrightarrow{e}_j$ if  $line(x_i(t), x_i^{des})$ intersects $line({x}_j(t) + S_{A} * \overrightarrow{e}_j, {x}_j(t) - \frac{S_A}{2} * \overrightarrow{e}_j)$, otherwise continues her free-flow motion. Here, $\overrightarrow{e}$ is the direction vector, $S_A$ is a scaling factor, $x(t)$ and $x_i^{des}$ are the current and final positions respectively. Cars continue by following their free-flow motion.
	\item Decelerate: Agents decelerate and in the end stop, if required. For pedestrians, $\text{newSpeed}_{i} = \frac{\text{Speed}_i(t)}{2}$, unless the car is very near (i.e., distance($i$, $j$) $\leq$ $r_i$ + $r_j$ + 1 m), in that case pedestrian will stop and in case of cars, $\text{newSpeed}_{j} = \text{Speed}_{j}(t) - \text{decRate}$.\newline
	\\
	Here, 
	$\text{decRate} = \begin{cases}
    \frac{\text{Speed}_{j}(t)}{2}, \text{if } \text{distance}(i, j) \leq D_{min}, \newline
    \\
    \frac{\text{Speed}_{j}^2}{\text{distance}(i, j) - D_{min}}, \text{otherwise}.
   \end{cases}$ 
   \newline
   $D_{min}$ is the critical spatial distance.
    \item Deviate: A pedestrian \textit{i} passes a car \textit{j} from behind from a position $p_i = {x}_j(t) - S_D*\overrightarrow{e}_j$ (up till \textit{j} stays within the range of view of \textit{i}) and afterwards \textit{i} proceeds moving towards her original goal position. 
\end{itemize}
Although these modules do not obey any sequence and take control alternatively, at the start of the simulation, GSFM keeps a hierarchy among them. It starts with trajectory planning, assuming that agents plan their trajectories before they begin moving. When trajectories are planned, the BDI controller actives the force-based module to model the physical movement of agents. Conflict recognition and classification are performed at regular intervals (the algorithm is given in \cite{johora2020zone}), and if it detects any complex conflict, then the controller activates the game-based module. As soon as the strategies are decided, the controller activates the force-based module again to execute the chosen strategies. The BDI controller also prioritizes different interactions based on their seriousness, for example, for cars, $\vec{I}_\text{stop}$ takes precedence over $\vec{I}_\text{game}$ and $\vec{I}_\text{game}$ obtains priority over car following. The following code fragment depicts the basic elements of a BDI program consisting of beliefs (in pink), plans (in blue), and actions (in black). 

\captionof{lstlisting}{Sample BDI Code Fragment}
\begin{lstlisting}[linewidth=\columnwidth,breaklines=true]
<@\textcolor{RubineRed}{module(3.0)}@>.
!<@\textcolor{blue}{main}@>.
+!<@\textcolor{blue}{main}@> <- 
    generic/print("Name", MyName);
    !!<@\textcolor{blue}{calculate/route}@>; !<@\textcolor{blue}{walk}@>.
    
+!<@\textcolor{blue}{calculate/route}@> <-
    route/calculate.
    
+!<@\textcolor{blue}{walk}@>: >>(<@\textcolor{RubineRed}{module(S)}@>, generic/type/isnumeric(S) && S==3.0) <- 
    calculate/next/position; !<@\textcolor{blue}{walk}@>.
    
+!<@\textcolor{blue}{update/belief(G)}@> <-
    >><@\textcolor{RubineRed}{module(S)}@>; -module(S); +module(G).
    
+!<@\textcolor{blue}{game/decelerate}@>: >> (<@\textcolor{RubineRed}{module(S)}@>, general/type/isnumeric(S) && S==1.0) <- 
    stop/moving; !<@\textcolor{blue}{game/decelerate}@>; !<@\textcolor{blue}{walk}@>.
\end{lstlisting}
Here, `+', `-', `$>>$' signify add (plan or belief), remove (belief) and unification (belief), respectively. The double exclamation mark before \textit{calculate/route} plan indicates that this plan should run in the current time-step and one exclamation mark before \textit{walk} says that the plan will execute in the next time-step. An agent can also trigger a plan from the environment. As an example, when the game module decides on the strategies for the road users involved in a conflict situation, it triggers the plan \textit{update/belief}, and the plan related to the decision, i.e., \textit{game/decelerate} in this sample (not complete) code fragment. 
\begin{equation}
\vspace{-3mm}
\text{Pedestrian:~}   
 \frac {d{\overrightarrow{v^t}_i}}{dt} = \Big(\overrightarrow{D}_{i}^o + \Sigma\overrightarrow{I}_{i W} + \Sigma\overrightarrow{I}_{ij} + w_p\cdot\overrightarrow{I}_{long} \Big) \hspace{0.02cm} or \overrightarrow{I}_{\text{game}}
 \label{eq:PedSFM}
\end{equation}
\begin{equation}
\text{Car:~}
 \frac {d{\overrightarrow{v^t}_i}}{dt} = \Big(\overrightarrow{D}_{i}^o + w_c\cdot\sum_{j\neq car}\overrightarrow{I}_{ij}\Big) \hspace{0.1cm} or \overrightarrow{I}_{\text{follow}} \hspace{0.1cm} or \overrightarrow{I}_{\text{game}} \hspace{0.1cm} or \overrightarrow{I}_{\text{stop}},
 \label{eq:CarSFM}
\end{equation}
\begin{equation}
 \hat{Y}^{t+\Delta t}_i = f \{ Z_i,  (\frac {d{\overrightarrow{v^t}_i}}{dt} + x_{i}(t))\}.
 \label{eq:modeling}
\end{equation}
The process of modeling the movements of any agent \textit{i} at any time step \textit{t} in GSFM is summarized in Eq.~\eqref{eq:PedSFM}--\eqref{eq:modeling}. Here, \textit{i}, \textit{j}, $W$, $Z_i$, $x_{i}(t)$, and $\hat{Y}^{t+\Delta t}_i$ denote the target agent, competitive agent, static obstacle, model inputs, the position of \textit{i} in current and next time step respectively. The input profile $Z_i$ contains start ($x^{st}_i$), goal ($x^{des}_i$), and speed profile of \textit{i}. The goal of \textit{i} is estimated by extending its last observed position ($x^{gt}_i$) in real trajectory using Eq.~\eqref{eq:destination} with the extended length $l_\text{des}$ = 5 m. The weight $w_p$ = 1 for the CITR scenarios, otherwise 0 and $w_c$ = 1 for the DUT scenarios, otherwise 0.
\begin{equation}
    x^{des}_i = x^{st}_i + l_\text{des}\cdot(x^{gt}_i - x^{st}_i),
    \label{eq:destination}
\end{equation}
We calculate the desired speed $v_\text{d}$ of a pedestrian by identifying the walking portion of his/her trajectory, i.e., where the pedestrian's speed is larger than a threshold $v_\text{walk}$ and then, we average all the speed values to obtain $v_\text{d}$. We set $v_\text{walk}=0.8 m/s$. A car's desired speed is set to: $mean(v_\text{i}) + std(v_\text{i}) * 0.5$, where $v_\text{i}$ is the set of all the speed values of car $i$.

\section{Calibration Methodology}
\label{se:calibration}
\begin{figure}[H]
\vspace{-4.5mm}
	\centering
	\includegraphics[width=3.2in]{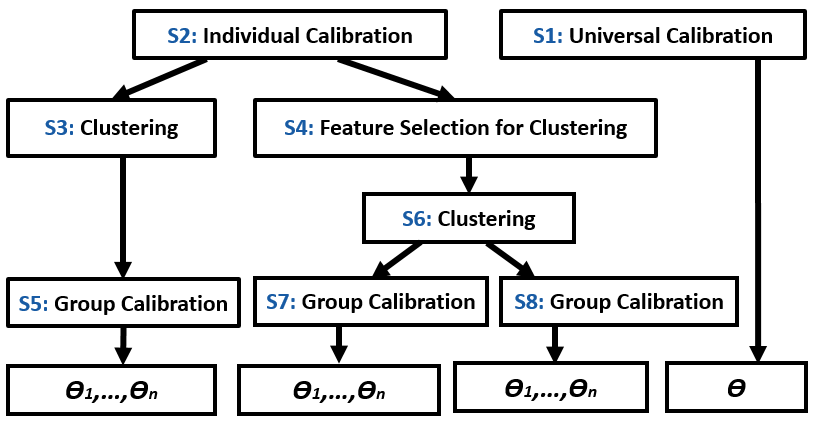}
	\caption{The workflow of model calibration.}
	\label{fig:cal_steps}
	\vspace{-3.5mm}
\end{figure}

\setlength{\parindent}{2ex}
In this paper, we calibrate our model parameters in several steps as visualized in Figure \ref{fig:cal_steps} and the calibration is performed using a genetic algorithm (see section \ref{se:genetic}). To recognize different motion patterns of pedestrians from real-world scenarios, we investigate two clustering approaches, namely Principal Component Analysis (\textbf{PCA}) with the k-means algorithm (step \textbf{S3}), and k-means with step-wise forward selection (\textbf{FS}) method (steps \textbf{S4} and \textbf{S6}), see section \ref{sc:clustering}. The steps in Figure \ref{fig:cal_steps} are as follows: We start by performing universal calibration to get one unique set of parameter values for all pedestrians by assuming that in the same situation, they all act similarly.

At the next step, we calibrate the parameters individually for each pedestrian, then cluster individual parameters using the above-mentioned clustering approaches which give us two different sets of pedestrian groups. Next, we perform group calibration (steps: \textbf{S5}, \textbf{S7} and \textbf{S8}) so that each group has a unique set of parameters values.
For the groups (i.e., clusters) that are obtained in step \textbf{S3}, we perform group calibration directly. However, for the groups obtained by completing \textbf{S4} and \textbf{S6}, we perform group calibration in two different phases i.e., \textbf{S7} and \textbf{S8}. In \textbf{S7}, we individually calibrate the selected parameters by the FS method for each group, while keeping the rest of the parameters' values (obtained in \textbf{S1}) same for all groups. Whereas in \textbf{S8}, we calibrate all parameters separately for each group. Each of these approaches above generates a different version of the GSFM model (see section \ref{se:genetic}).

GSFM contains a large set of parameters, which can be broadly classified into parameters for SFM interaction, safety measurements, and payoff matrix calculation for game playing. The SFM and safety-related parameters are listed in Table~\ref{tab:sfmandsafetyparam} and Table \ref{table:gameparameters} shows the game parameters. Among these parameters, for grouping pedestrians, we select the sets of parameters given in Table \ref{table:clusteringparameters} based on sensitivity analysis. The rest of the parameters are calibrated universally as step \textbf{S1}.
\begin{table}[htbp]
   \vspace{-2.5mm}
	\centering
	\setlength{\tabcolsep}{9.5pt}
	\renewcommand{\arraystretch}{1.5}
	\caption{The list of parameters calibrated for clustering}
	\begin{tabular}{ |p{8cm}| }
		\hline
		 Interaction strength: \large $V_{ij}^o$ \small(PP), \large$V_{ij}^o$ \small(PC), \large$V_{ij}^o$ \small(CP), \\
		 Repulsive interaction range: \large $\sigma$ \small(PP), \large $\sigma$ \small(PC),\\ Anisotropic parameter: \large$\lambda$, \small Scaling factor for deviate action: \large$S_D$ \\
		\hline
	\end{tabular}
	\label{table:clusteringparameters}
	\vspace{-4.5mm}
\end{table}

\vspace{-2mm}
\subsection{Clustering}
\label{sc:clustering}
\paragraph{K-means with Principal Component Analysis} \textbf{K-means} is a simple, fast and widely used clustering algorithm for classifying data based on euclidean distance between the data points, with a predefined number of clusters \cite{marutho2018determination}. In this paper, we decide on the number of clusters using the elbow method \cite{marutho2018determination}, and each data point represents the calibrated parameters' values of an individual pedestrian. 

\textbf{Principal Component Analysis} \cite{marutho2018determination} is a technique that reduces a larger number of parameters to a smaller set of parameters which are linear combinations of the original parameters and contains most of their information. As stated in \cite{ding2004k}, reducing the dimension of data using PCA is beneficial for k-means. Thus, we use PCA to reduce the number of parameters given in Table \ref{table:clusteringparameters}, and then perform k-means on the reduced parameters set to cluster pedestrians into groups.

\paragraph{K-means with Forward Selection}
\textbf{Forward selection} is a simple but commonly used feature (or parameter) selection method. It starts with a empty model which contains no parameters, then continue adding the most significant parameter one after another until a predefined stopping criteria has reached or if all present parameters are already in the model \cite{borboudakis2019forward}.

\begin{algorithm}
\KwInput{Number of clusters $K$, Set of parameters $A_p$, Predefined score $C_s$}
\KwOutput{Set of selected parameters for clustering $S_p$}
    $S_p \gets  \{\}$\;
    $Cs_{int}$ = 0\tcp*{initialize clustering score to 0}
    $temp_s = 0$\; $temp_p \gets \{\}$\;
    \While{$Cs_{int} < C_s$}
    {
       \For{each $p$ $\in$ \vars{$A_p$}}
        {
           \If{$S_p$ == $\emptyset$}
           {
              perform k-means clustering for $p$\;
           }
           \Else
           {  
               $t_p$ = $S_p \cup \{p\}$\;
               perform k-means clustering for $t_p$\;
           }
           $score$ = silhouette score of clusters\;
           \If{$temp\_s < score$}
           { 
             $temp_s = score$\;
             $temp_p = p$\;
           }
        }
        $S_p$ = $S_p \cup \{temp_p\}$\;
        $A_p$ = $A_p \setminus \{temp_p\}$\;
        $Cs_{int} = temp_p$\;
    }
    \caption{Forward Selection with k-means}
    \label{al:simulation}
     \vspace{-1.5mm}
\end{algorithm}
\begin{figure}[H]
\vspace{-2.5mm}
\begin{minipage}{1\columnwidth}
\includegraphics[width=0.495\textwidth]{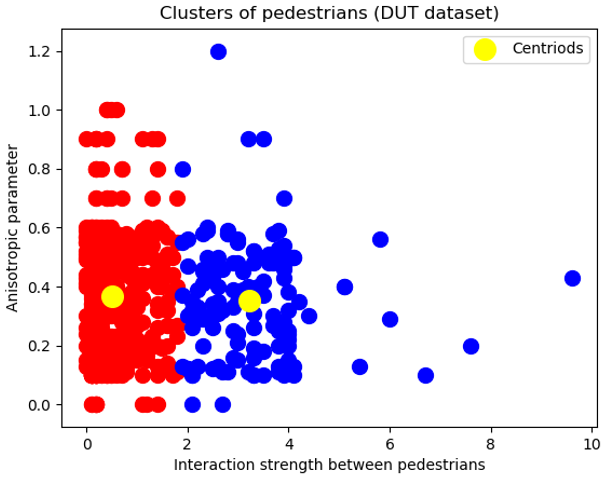}
\includegraphics[width=0.495\textwidth]{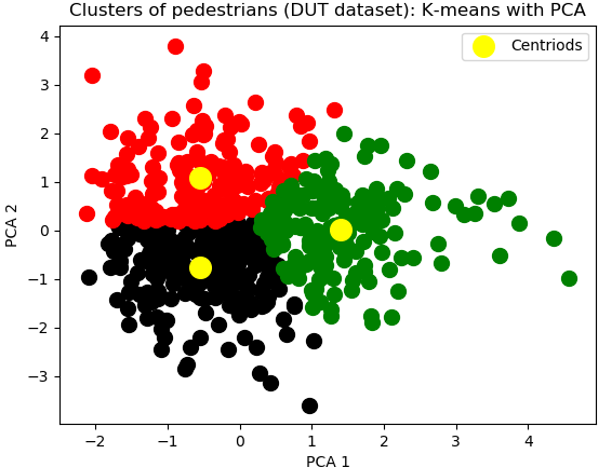}
\end{minipage}
\hfill
   \caption{Different pedestrians groups of the DUT data set with different motion patterns.}
  \label{fig:clusters}
  \vspace{-2.5mm}
\end{figure}
We calculate the significance of the parameter(s) by executing k-means for some k (i.e., number of clusters) and measure the clustering performance using the silhouette score. This method terminates if a preset value of silhouette score has been reached. The silhouette value is a measure to see if a data point is similar to its own cluster than to others \cite{ROUSSEEUW198753}. Algorithm \ref{al:simulation} shows the steps of the forward selection method with k-means. After performing feature selection using Algorithm \ref{al:simulation}, we perform k-means on the reduced set of parameters to cluster pedestrians into groups with different motion patterns.

Figure \ref{fig:clusters} shows different clusters of pedestrians from the DUT data set obtained by performing k-means with forward selection and k-means with PCA, from left to right. We conduct these approaches separately on each data set.

\vspace{-2mm}
\subsection{Calibration}
\label{se:genetic}
Genetic algorithms (\textbf{GA}) \cite{zames1981genetic} are evolutionary algorithms, largely applied to tackle optimization problems such as calibration of model parameters \cite{amirjamshidi2019multi,schiermeyer2016genetic}. 

As stated earlier, we calibrate our model parameters using a GA. It begins with feeding a random initial set of chromosomes i.e., the set of parameters that need to be calibrated into the simulation model to acquire and compare outputs with real-world data to compute and assign a fitness score to the respective chromosome. Next, an offspring population is generated by performing the selection (of the fittest members), crossover, and mutation processes and fed into the model again unless a specific stopping criterion has reached. 

We only consider the parameters in Table \ref{table:clusteringparameters} for grouping pedestrians, and we calibrate these parameters as illustrated in Figure \ref{fig:cal_steps}. Whereas, we calibrate the rest of the parameters of GSFM in beforehand, separately and in two steps: first, we calibrate the remaining SFM and safety parameters and then calibrate the game parameters. We conduct all these calibration steps using the above-described genetic algorithm. To be noted, during individual calibration of pedestrian, we simulate only the target pedestrian and update the states of surrounding agents as their real trajectories.

Selection of the fitness function and simulation output type depends on the types of parameters to calibrate. To calibrate the SFM and safety parameters, GSFM outputs the simulated positions of agent(s) ($\vec{P}^{sim}_u$) to compare with their real positions ($\vec{P}^{real}_u$) for calculating the fitness score of any respective chromosome. For the universal and group calibration, the fitness score is calculated by Eq.\ref{eq:sfmfitnessfunction} and the fitness function for the individual calibration is given in Eq.\ref{eq:individualfitness}.
 \vspace{-1.5mm}
\begin{equation}
   f_{score} = 
   \Bigg( \sum^{E}_\textit{e} \bigg(\sum^{U}_\textit{u} 
   \Big( \sum^{T}_\textit{t}\Bigl\lvert\vec{P}^{real}_u (\textit{t}) \hspace{0.02cm} - \hspace{0.02cm} \vec{P}^{sim}_u (\textit{t})\Bigr\rvert\Big)\big/T \bigg) \Big/ U \Bigg) \bigg/E
   \label{eq:sfmfitnessfunction}
\end{equation}
\begin{equation}
   f_{score} = 
   \Big(\sum^{T}_\textit{t}\Bigl\lvert\vec{P}^{real}_u (\textit{t}) \hspace{0.02cm} - \hspace{0.02cm} \vec{P}^{sim}_u (\textit{t})\Bigr\rvert\Big)\big/T
   \label{eq:individualfitness}
\end{equation}
\begin{equation}
f_{score} = 
\sum^{E}_\textit{e} \Bigg(
\bigg(\sum^{U}_\textit{u}
\begin{cases}
1, & if~A^{real}_u \hspace{0.02cm} == \hspace{0.02cm} A^{sim}_u\\
-1, & \text{otherwise}
\end{cases}
\bigg)
\bigg/ U \Bigg) 
\Bigg/ E
\label{eq:gamefitnessfuction}
\end{equation}
Here, $E$, $U$, and $T$ denote the number of scenarios, the number of agents, and the number of time steps, respectively.
For Eq. \ref{eq:gamefitnessfuction}, the simulated decisions ($A^{sim}_u$) are obtained by game playing and the real decisions ($ A^{real}_u$) are manually extracted from the video data. To calibrate the game parameters, calculating the fitness score using Eq. \ref{eq:gamefitnessfuction} is preferable, as the game module is responsible for deciding on decisions/strategies for agents in any conflict situation, not their motion (see Section \ref{se:model}). We use Eq. \ref{eq:gamefitnessfuction} for calibrating the game parameters for the HBS data set but in case of the CITR and DUT data sets, Eq. \ref{eq:sfmfitnessfunction} is used due to the difficulty on extracting the real decisions manually.

The values of the game parameters are given in Table \ref{table:gameparameters}. Table \ref{tab:sfmandsafetyparam} shows the values of the SFM and safety-related parameters with their calibrated values, where, PP, PC, CP, and CC denote pedestrian-to-pedestrian,  pedestrian-to-car, car-to-pedestrian, and car-to-car interactions, respectively.
\begin{table}[htbp]
\vspace{-3mm}
	\centering
	\setlength{\tabcolsep}{8.5pt}
	\renewcommand{\arraystretch}{1.5}
	\caption{List of game parameters with calibrated values}
	\begin{tabular}{|c|c|c|c|}
		\hline
		\textbf{Symbol} & \textbf{HBS Value} & \textbf{DUT Value} & \textbf{CITR Value}\\
		\hline
		$G^C_{speed}$   & 11 & 4 & 10.4  \\
		\hline
		$G^P_{speed}$   & 1 & 0 & 1  \\
		\hline
		$G^{competitor}_{speed}$ & 11 & 0 & 6.3  \\
		\hline
		$G_{noai}$   & 3 & 0  & 0.3  \\
		\hline
		$G_{stopped}$  & 2 & 0 & 1.1  \\
		\hline
		$G^F_{angle}$ & 1 & 6.6 & 0.4 \\
		\hline
		$G^{min}_{dis}$ & 7 & 5 & 6.1 \\
		\hline
		$G^{Ace}_{angle}$ & 7 & 8 & 7 \\
		\hline
		$G^{Dec}_{angle}$ & 5 & 8 & 5 \\
		\hline
		$G^{Dev}_{angle}$ & 8 & 6 & 8 \\
		\hline
	\end{tabular}
	\label{table:gameparameters}
	\vspace{-2mm}
\end{table}

\begin{table*}[htbp]
    \centering
    \caption{The list of the SFM and safety parameters with their calibrated values. Here, G1, G2, G3 are the clustered groups.}
    \begin{tabular}{|l|l|l|l|l|l|l|l|l|l|l|l|l|l|}
\hline
\textbf{Symbol} & \textbf{Description} & \textbf{Unit} &\multicolumn{4}{l|}{\textbf{HBS}}&\multicolumn{3}{l|}{\textbf{DUT}}&\multicolumn{4}{l|}{\textbf{CITR}}\\
\cline{4-14}
 &&&\textbf{G1}&\textbf{G2}&\textbf{G3}&\textbf{U}&\textbf{G1}&\textbf{G2}&\textbf{U}&\textbf{G1}&\textbf{G2}&\textbf{G3}&\textbf{U}\\
\hline
$V_{ij}^o$($PP$)  &Interaction strength &  $m^2s^{-2}$ & 0.1 & 0.1 & 1.9& 0.1& 0.01&0.1 &0.1 &0.2& 0.1 & 0.4 &0.1\\
 \hline
$V_{ij}^o$($PC$)  &Interaction strength &  $m^2s^{-2}$ & 15.1 &17.3 & 11.9& 11.7&1.6 &3.4 & 4.5& 0.2 & 2.6&0.07&1.5 \\
 \hline
$V_{ij}^o$($CP$)  &Interaction strength &  $m^2s^{-2}$ & --- &--- & ---& ---&0.76 & 1.7& 2.27& --- & ---&---&--- \\
  \hline
 $\sigma$ ($PP$) & Repulsive interaction range &$m$ & 0.17 &0.24 &0.25 & 0.25&0.17 &0.18 & 0.23& 0.1&0.2&0.25&0.18\\
   \hline
 $\sigma$ ($PC$) &  Repulsive interaction range  &$m$ & 0.1 & 0.7 &0.2 &0.91 & 0.11&0.14 & 0.27&1.5 &0.39&1.1&0.69\\
  \hline
 $\lambda$ &  Anisotropic parameter & --- & 0.35& 0.339 &0.42 &0.35 & 0.43& 0.16& 0.41&0.15&0.59&0.52&0.13\\
 \hline
 $S_D$ & Scaling factor for deviate action&--- & 7.6& 12& 7.8 &6 & 6& 7& 9.01&6.1&8.4&8.3&7.0\\
  \hline
$V_R$ & Range of view   &$m$ &\multicolumn{4}{l|}{18.4}&\multicolumn{3}{l|}
 {10}&\multicolumn{4}{l|}{12.3}\\
 \hline
 $D_{min}$($PC$) & Critical spatial distance &$m$ &\multicolumn{4}{l|}{7.8}&\multicolumn{3}{l|}
 {8}&\multicolumn{4}{l|}{7}\\
\hline
 $S_A$  & Scaling factor for accelerate action   &---& \multicolumn{11}{l|}{6}\\
\hline
 $S_C$ & Scaling factor for conflict detection &---& \multicolumn{11}{l|}{9} \\
\hline
 $D_{min}$($CC$)   & Critical spatial distance   &$m$ & \multicolumn{11}{l|}{8}\\
\hline
 $U_{i B}^o$  & Interaction strength for obstacle   &$m^2s^{-2}$ & \multicolumn{11}{l|}{10}\\
 \hline
 $\gamma$ (obstacle)  & Repulsive interaction range  &$m$ & \multicolumn{11}{l|}{0.2}\\
 \hline
\end{tabular}
\label{tab:sfmandsafetyparam}
\vspace{-2mm}
\end{table*}

\begin{table*}[htbp]
    \centering
    \caption{Quantitative results i.e., \textbf{aADE}(m) / \textbf{aFDE}(m) / \textbf{SD}($ms^{-1}$) / \textbf{CI} of the classical SFM and all versions of GSFM. Here, the {\bf bold} number denotes the best score.}
    \begin{tabular}{|l|l|l|l|l|l|l|}
\hline
\textbf{Model} &\multicolumn{3}{l|}{\textbf{pedestrian}}&\multicolumn{3}{l|}{\textbf{Vehicle}}\\
\cline{2-7}
 &\textbf{HBS}&\textbf{DUT}&\textbf{CITR}&\textbf{HBS}&\textbf{DUT}&\textbf{CITR}\\
\hline
GSFM-M1& \textbf{0.745}/\textbf{0.807}/0.338/0.0182& 0.654/1.07/0.261/0.033& 0.565/0.859/0.1742/0.0037& \textbf{1.26}/\textbf{3.33}/\textbf{1.083} & 1.29/3.33/0.795& \textbf{2.41}/\textbf{5.17}/\textbf{1.153}\\ 
\hline
GSFM-M2&0.747/0.812/\textbf{0.333}/\textbf{0.0112}& \textbf{0.643}/\textbf{1.06}/0.263/0.036& \textbf{0.546}/\textbf{0.813}/0.1754/0.0037& 1.33/3.46/1.107 & \textbf{1.22}/\textbf{3.04}/\textbf{0.787}& 2.46/5.19/1.166 \\ 
\hline
GSFM-M3&0.766/0.854/0.338/0.0138&
0.698/1.19/\textbf{0.260}/0.033&
0.577/0.878/0.1742/\textbf{0.0035}&
1.28/3.39/1.094&
1.25/3.27/0.803&
2.49/5.28/1.183\\ 
\hline
GSFM-U&0.754/0.829/0.335/0.0127&
0.705/1.22/0.265/\textbf{0.030}&
0.577/0.880/\textbf{0.1740}/0.0035&
1.30/3.42/1.097&
1.41/3.51/0.842&
2.49/5.29/1.180\\ 
\hline
SFM&1.122/1.164/0.376/0.0305&
1.499/2.26/0.263/0.036&
1.185/1.791/0.2566/0.0123&---&---&---
 \\ 
\hline
\end{tabular}
\label{tab:results}
\vspace{-3mm}
\end{table*}

After performing the clustering and calibration processes, we got several sets of parameters which results in different versions of our model. Specifically, GSFM-M1 which indicates the model with k-means and PCA, GSFM-M2 is the model that combines the forward selection method with k-means and calibrates all parameters given in Table \ref{table:clusteringparameters} during group calibration (\textbf{S8}), GSFM-M3 is the model with FS and k-means where only the selected parameters by FS are calibrated in group calibration (\textbf{S7}), and GSFM-U denotes the universal model, i.e., the model with one set of parameters. Due to space restrictions, Table \ref{tab:sfmandsafetyparam} visualizes only the the values of parameters in GSFM-U (denoted as \textbf{U}) and GSFM-M2, for each data set. Here, G1, G2, G3 denote the clusters or groups.

\vspace{-2mm}
\begin{figure*}
	\centering
	\includegraphics[width=5.85in,height=3.8in]{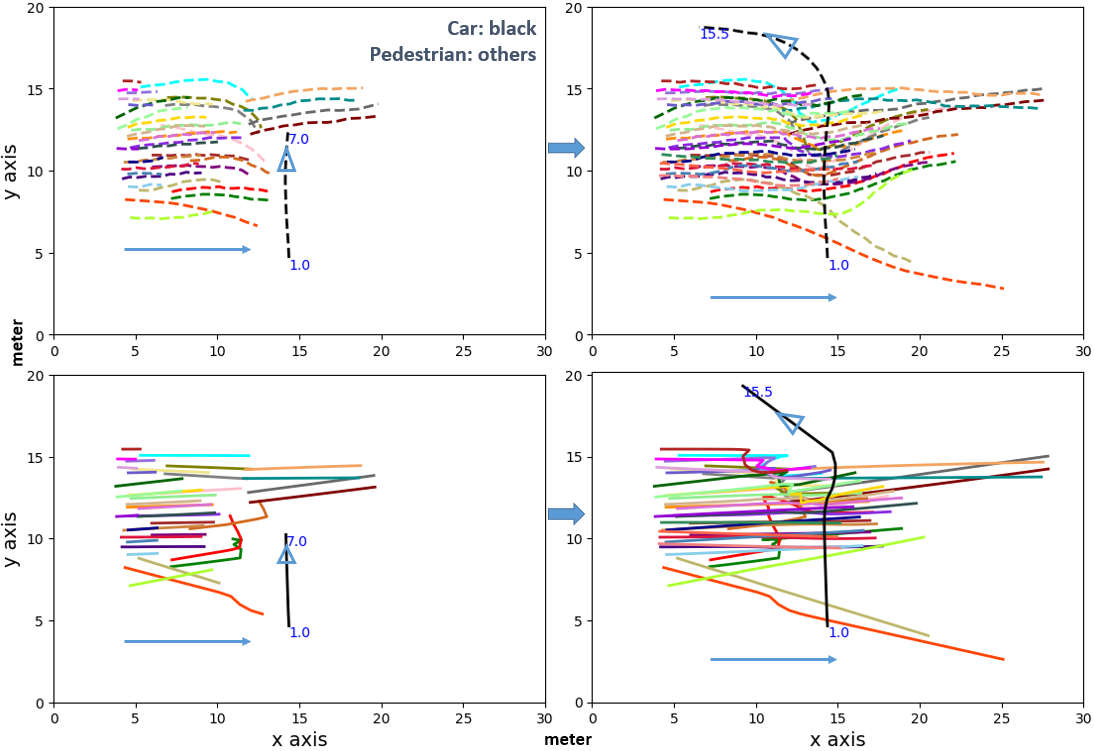}
	\caption{Crowd-to-car interaction from the DUT data set. The first row shows the real trajectories and the second row depicts the simulated trajectories, at two subsequent time steps. Car's trajectories are in black color.}
	\label{fig:evscene1}
	\vspace{-3.5mm}
\end{figure*}

\begin{figure*}
	\centering
	\includegraphics[width = 7.0in]{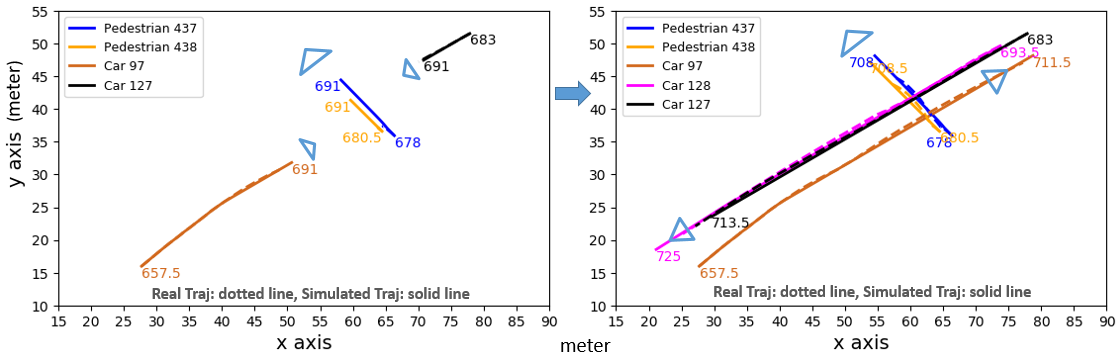}
	\caption{Pedestrians-to-cars crossing scenario from the HBS data set. The dotted lines represent the real trajectories and the solid lines are the simulated trajectories. Trajectories are visualized at two subsequent time steps.}
	\label{fig:evscene2}
	\vspace{-5mm}
\end{figure*}

\section{Evaluation}
\label{se:evaluation}
As a quantitative evaluation, we compare all our models, namely GSFM-M1, GSFM-M2, GSFM-M3 and GSFM-U and the classical SFM proposed in~\cite{helbing2000simulating}. We calibrate all parameters of the classical SFM for each data set using the GA in Section \ref{se:genetic} and the fitness function in Eq. \eqref{eq:sfmfitnessfunction}, for a fair comparison. The performances of these models are evaluated by the metrics given in Section \ref{sec:evalmetrics} on the extracted interaction scenarios from the HBS, DUT and CITR data sets (summarized in Table \ref{tab:data_statistics}). We select three example scenarios among all to evaluate the performance of our model qualitatively. We run all simulations on an Intel$\circledR$ Core\texttrademark i5 processor with 16 GB RAM.

\vspace{-3.5mm}
\subsection{Evaluation Metrics}
\label{sec:evalmetrics}
To evaluate the performance of the proposed models in terms of how realistic the resulting trajectories are, we consider two most commonly used metrics \cite{rudenko2020human,alahi2017learning}, namely average displacement error (ADE) and final displacement error (FDE), together with two other metrics:
\begin{itemize}
    \item \textbf{Adjusted Average Displacement Error (aADE)}: ADE computes the pairwise mean square error (in meter m) between the simulated and real trajectories of each agent over all positions and averages the error over all agents. In our extracted scenarios, the trajectory length of agents $k$ are different; thus, we choose an adjusted version of ADE to evaluate our models' performance more precisely: $\text{aADE}=\frac{k_0}{k}\text{ADE}$, with $k_0$ as a predefined trajectory length (i.e., number of time steps), assuming that the error in trajectory modeling increases linearly.
    \item \textbf{Adjusted Final Displacement Error (aFDE)}: FDE calculates the average displacement error (in m) of the final point of all agents. We also adjust FDE like $\text{aADE}$. 
    \item \textbf{Speed Deviation (SD):} the SD metric is for measuring the pairwise speed difference (in $ms^{-1}$) of simulated and real speed of each agent over all time steps and averaging these difference over all agents. \text{SD} is adjusted as \text{aADE}.
    \item \textbf{Collision Index (CI):} We choose the \text{CI} metric to penalize any collision of pedestrian(s) with the car(s). For each pedestrian $i$, $\text{CI}\in [0,1]$ is described as the portion of the simulated trajectory of $i$ that overlaps with any car's occupancy. $\text{CI}=0$ means no collision. \text{CI} is averaged over all pedestrians and adjusted as other metrics.
\end{itemize}

\vspace{-5mm}
\subsection{Results}
\setlength{\parindent}{2ex}
Table \ref{tab:results} visualizes the performances of the GSFM-M1, GSFM-M2, GSFM-M3, GSFM-U and the classical SFM models on the HBS, DUT and CITR data sets, evaluated using the above-described metrics. In column entries of Table \ref{tab:results}, for pedestrians, we reported four scores that are aADE, aFDE, SD, and CI, respectively and for cars, three scores are shown as CI is only calculated from the perspective of pedestrians. The bold number indicates the best score. In all criterion, the GSFM-M1 and GSFM-M2 models perform similarly, and both these models outperform the universal model GSFM-U, but GSFM-M3 performs mostly similar to GSFM-U. All versions of GSFM model always perform better than the classical SFM. For all data sets, the average errors of our best-performed model in trajectory modeling, i.e. aADE and aFDE is range from 0.5 m to 1 m for pedestrian, which considers as a good result given the stochasticity in pedestrians behaviors and also similarities with the results presented in \cite{sadeghian2019sophie}, a state-of-the-art trajectory prediction model of pedestrians that evaluated by pedestrian-only scenarios. However, the aADE/aFDE scores of our model for vehicles is comparatively higher than pedestrians, i.e. bigger error, mainly for the CITR data set. One reason behind this is the significant difference in simulated and real speeds of vehicles. Thus, improving our vehicle motion modeling, e.g., by considering different motion patterns and speed profiles of vehicles, is part of our future work.

In all cases, the collision index CI is minimal, which indicates all models simulate collision-free trajectories for most of the time. Moreover, in terms of CI, our models perform much better than SFM for the CITR and HBS data sets, but due to higher pedestrian density in DUT, the performance of our models drop and become similar to SFM. For SFM, the entries for cars are empty because the classical SFM can only model pedestrian motions. Thus, in SFM, during the simulation of the extracted scenarios, the cars follow their real trajectories.  
\begin{figure*}[htbp]
\vspace{-1.5mm}
	\centering
	\includegraphics[width = 7.1in]{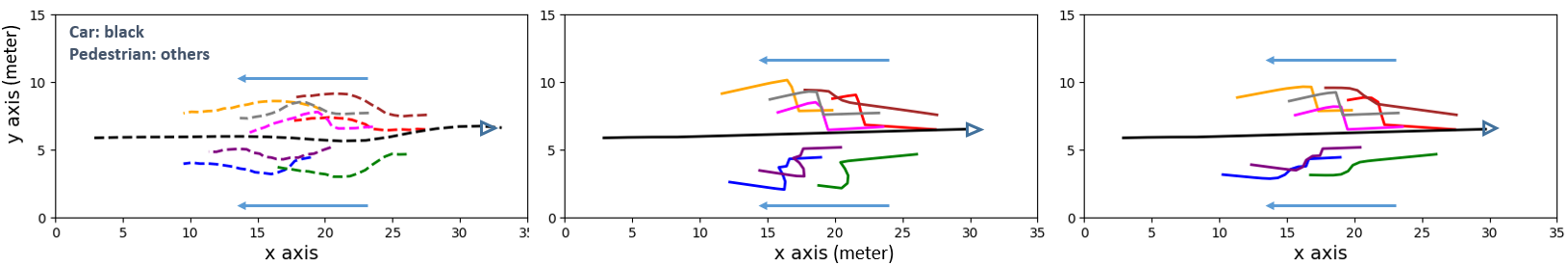}
	\caption{Pedestrians-to-car interaction from the CITR data set. The trajectories of road users: real, simulated in GSFM-U, and simulated in GSFM-M2 are visualized respectively, from left to right.}
	\label{fig:evscene3}
	\vspace{-5mm}
\end{figure*}

To show the differences in the DUT, HBS and CITR data sets and the capability of our model to address these differences, we choose one scenario from each data set and simulate each scenario in GSFM-M2. In all Figures \ref{fig:evscene1}, \ref{fig:evscene2}, and \ref{fig:evscene3}, the dotted lines indicate the real trajectory and the solid lines represent the simulated trajectories of road users. In Figure \ref{fig:evscene1} and Figure \ref{fig:evscene2}, the real and simulated trajectories are visualized at two specific subsequent time steps. The black lines in Figure \ref{fig:evscene1} and Figure \ref{fig:evscene3} indicate the trajectories of car and the color-coded lines depict the trajectories of pedestrians. 

Figure \ref{fig:evscene1} visualizes a crowd-to-car interaction scenario from the DUT data set. Here, the first row shows the real trajectories of the involved road users, and the second row visualizes the simulated trajectories. Most of the DUT scenarios contain a large number of pedestrians, as shown in Figure \ref{fig:evscene1}.

Figure \ref{fig:evscene2} depicts a complex pedestrians road crossing example with cars coming from two directions, extracted from the HBS data set. Both in simulation and reality, both cars stop to let the pedestrians cross first, which is a common phenomenon in HBS scenarios. 

Figure \ref{fig:evscene3} shows a pedestrians-to-car interaction scenario from CITR. As visualized in Figure \ref{fig:evscene3}, GSFM-U simulates all pedestrians in a similar style, while in GSFM-M2, pedestrians follow different motion patterns. Thus, the simulated trajectories of pedestrians in GSFM-M2 are more identical to their real trajectories than the trajectories generated by GSFM-U.

To sum up, in all example scenarios, our model realistically simulates complex interactions among pedestrians and car(s). Table \ref{tab:results} shows that our model performs satisfactorily for all data sets. Thus, our model was able to model scenarios from new data sets convincingly (i.e. CITR and DUT) with minimal effort compared to traditional approaches (i.e. starting modeling process from scratch for each new case), through the integration of new types of interactions into the model and largely automated calibration process. This evaluates the generalizability of our model. Plus, the results of our quantitative evaluation and the visualization and discussion of the scenario in Figure \ref{fig:evscene3} state that the performance of our model is improved due to heterogeneous motion patterns of pedestrians.

 \vspace{-1.5mm}
\section{Conclusion and Future Work}
\label{sec:conclusion}
In this paper, we proposed a procedure to formulate general motion models and applied this process to extend our Game-Theoretic Social Force Model (GSFM) towards a general model for generating realistic trajectories of pedestrians and cars in different shared spaces. Secondly, we applied and examined two clustering approaches namely, Principal Component Analysis (PCA) with the k-means algorithm and k-means with the forward selection method, to recognize and model different motion patterns of pedestrians.

We calibrated, validated, and evaluated our model using three shared space data sets, namely the HBS, DUT and CITR data sets. These data sets differ from one another in terms of spatial layout, types of interactions, traffic culture and density. In both quantitative and qualitative evaluation process, our model performed satisfactorily for each data set, which evinces that by following a systematic procedure with a well-defined calibration methodology, a shared-space model can adapt to a new environment and model a large variety of interactions. The results also indicate that the heterogeneity in pedestrians motion improves the performance of our model.

Our future research will focus on improving the motion model for vehicles, adding new modalities (e.g., cyclists) into our model, calibrating the model parameters for a wider range of interactions (e.g., vehicle-to-vehicle complex interaction), recognizing different motion patterns of other user types such as vehicles, and calibrating and evaluating our model using more open-source data sets of shared spaces. Most significantly, we shall study large scenarios with a larger number of participants to investigate the scalability of different interaction types and also our simulation model.

\appendices

\vspace{-3mm}
\section*{Acknowledgment}
This work is supported by the German Research Foundation (DFG) through the Research Training Group SocialCars (GRK 1931) and by the United States Department of Transportation under (\#69A3551747111) for the Mobility21 University Transportation Center. We acknowledge the DFG research project MODIS (\#248905318) for sharing the HBS data set.
\vspace{-3mm}

\ifCLASSOPTIONcaptionsoff
  \newpage
\fi



\bibliographystyle{IEEEtran}
\bibliography{mybib}



\begin{IEEEbiography}[{\includegraphics[width=1in,height=1.25in,clip,keepaspectratio]{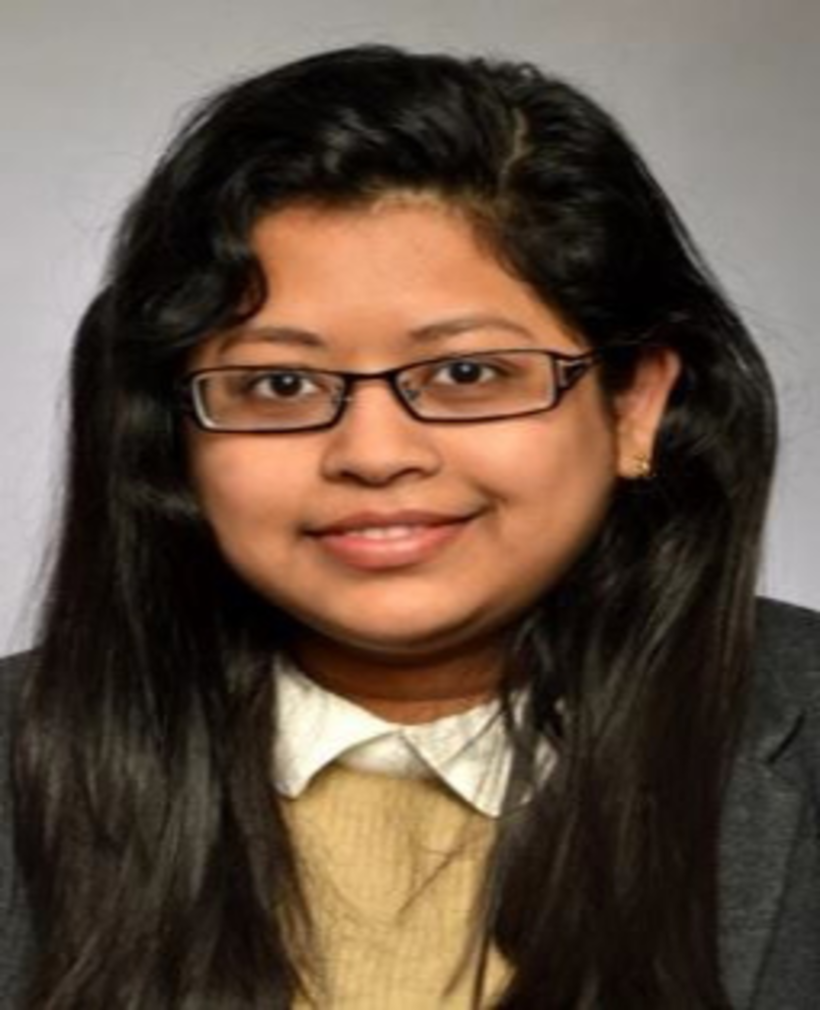}}]{Fatema T. Johora} received her B.Sc. degree in Computer Science and Engineering from Jessore University of Science and Technology, Bangladesh, in 2013, and her M.Sc. degree in Internet Technologies and Information Systems from Clausthal University of Technology, Germany, in 2017. She is currently a doctoral candidate at the Department of Informatics at Clausthal University of Technology. 

Her research interests cover game theory, agent-based modeling, and machine learning in the area of intelligent transport system and autonomous driving.
\end{IEEEbiography}

\begin{IEEEbiography}[{\includegraphics[width=1in,height=1.25in,clip,keepaspectratio]{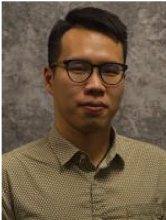}}]{Dongfang Yang}

Dongfang Yang received his bachelor's degree in microelectronics from Sun Yat-sen University, Guangzhou, China, in 2014. He has been with The Ohio State University since 2015 and received his Ph.D. in Electrical and Computer Engineering from The Ohio State University, in 2020. He is currently a graduate research associate at The Ohio State University. His research interests include control systems, computer vision, and machine learning with applications in intelligent transportation and autonomous driving.
\end{IEEEbiography}

\begin{IEEEbiography}[{\includegraphics[width=1in,height=1.25in,clip,keepaspectratio]{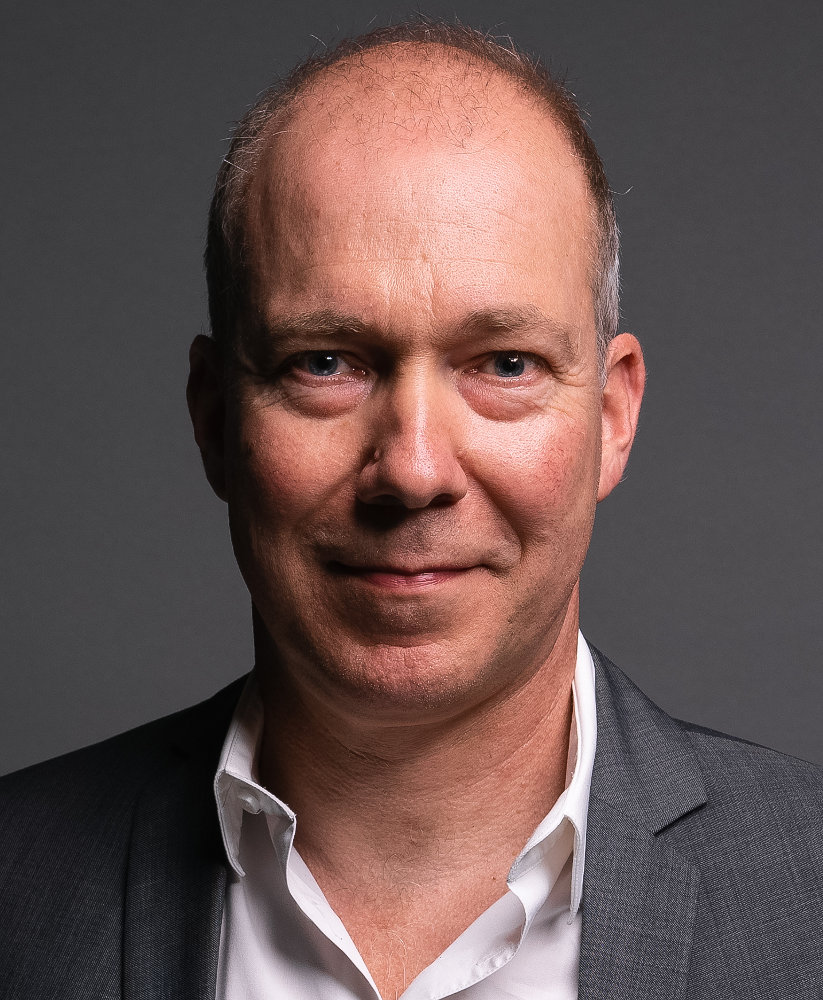}}]{Jörg P. Müller} is a Full Professor of Computer Science at the Department of Informatics at Clausthal University of Technology. He holds a Ph.D. in Computer Science from Universität des Saarlandes in the area of Artificial Intelligence, Intelligent Agents and Multiagent Systems. Prior to becoming a professor, he obtained ten years of industrial research experience in agent technology and peer-to-peer computing, working for Mitsubishi Electric, John Wiley and Sons, and Siemens Corporate Technology. His current research interests cover the broad area of modelling and simulation of socio-technical systems, coordination and intelligent systems. A long-term research focus is on agent-based modelling and simulation in the area of intelligent transport systems and future connected traffic systems.  Jörg has served on numerous conference committees in the area of AI, intelligent agents and multi-agent systems and has co-authored over 250 scientific publications.
\end{IEEEbiography}

\begin{IEEEbiography}[{\includegraphics[width=1in,height=1.25in,clip,keepaspectratio]{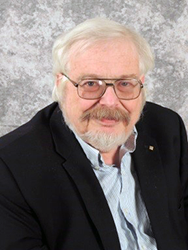}}]{Ümit Özgüner} (S’72–M’75–F’10) Prof. Emeritus Ümit Özgüner, TRC Inc. Chair on ITS at The Ohio State University, is a well know expert on Intelligent Vehicles. He holds the title of “Fellow” in IEEE for his contributions to the theory and practice of autonomous ground vehicles and is the Editor in Chief of the IEEE ITS Society, Transactions on Intelligent Vehicles.

He has led and participated in many autonomous ground vehicle related programs like DoT FHWA Demo’97, DARPA Grand Challenges and the DARPA Urban Challenge. His research has been (and is) supported by many industries including Ford, GM, Honda and Renault. He has published extensively on control design and vehicle autonomy and has co-authored a book on Ground Vehicle Autonomy. His present projects are on Machine Learning for driving, pedestrian modeling at OSU and participates externally on V\&V and risk mitigation, and self-driving operation of specialized vehicles. Professor Ozguner has developed and taught a course on Ground Vehicle Autonomy for many years and has advised over 35 students during their studies towards a PhD.
\end{IEEEbiography}

\end{document}